\documentclass[10pt,twocolumn,letterpaper]{article}

\usepackage{cvpr}
\usepackage{times}
\usepackage{epsfig}
\usepackage{graphicx}
\usepackage{amsmath}
\usepackage{amssymb}

\usepackage[bookmarks=false]{hyperref}

\cvprfinalcopy 

\def\cvprPaperID{2370} 

\begin{document}

\title{Connecting Look and Feel: Associating the visual and tactile properties of physical materials}

\author{Wenzhen Yuan$ ^1$\thanks{Equal Contribution} , Shaoxiong Wang$^{2,1}$\footnotemark[1] , Siyuan Dong$^1$, Edward Adelson$^1$ \\
	$^1$ MIT, $^2$ Tsinghua University \\
{\tt\small \{yuan\_wz, wang\_sx, sydong, eadelson\}@mit.edu}
}

\maketitle

\begin{abstract}
	
	For machines to interact with the physical world, they must understand the physical properties of objects and materials they encounter. We use fabrics as an example of a deformable material with a rich set of mechanical properties. A thin flexible fabric, when draped, tends to look different from a heavy stiff fabric. It also feels different when touched. Using a collection of 118 fabric sample, we captured color and depth images of draped fabrics along with tactile data from a high-resolution touch sensor. We then sought to associate the information from vision and touch by jointly training CNN’s across the three modalities. Through the CNN, each input, regardless of the modality, generates an embedding vector that records the fabric's physical property. 
	By comparing the embeddings, our system is able to look at a fabric image and predict how it will feel, and vice versa. 
	We also show that a system jointly trained on vision and touch data can outperform a similar system trained only on visual data when tested purely with visual inputs.

\end{abstract}

\section{Introduction}

The success of computer vision has been greatly accelerated through the use of deep learning and convolutional neural networks (CNN). However, the main successes have been with passive tasks; for example, the computer is given an image and in response it provides categories or descriptors.  For a machine to more actively interact with objects in the physical world, it must understand something about their physical properties. Vision can be used, for example, to predict how an object will feel when touched.

\begin{figure}
	\centering{
		\includegraphics[]{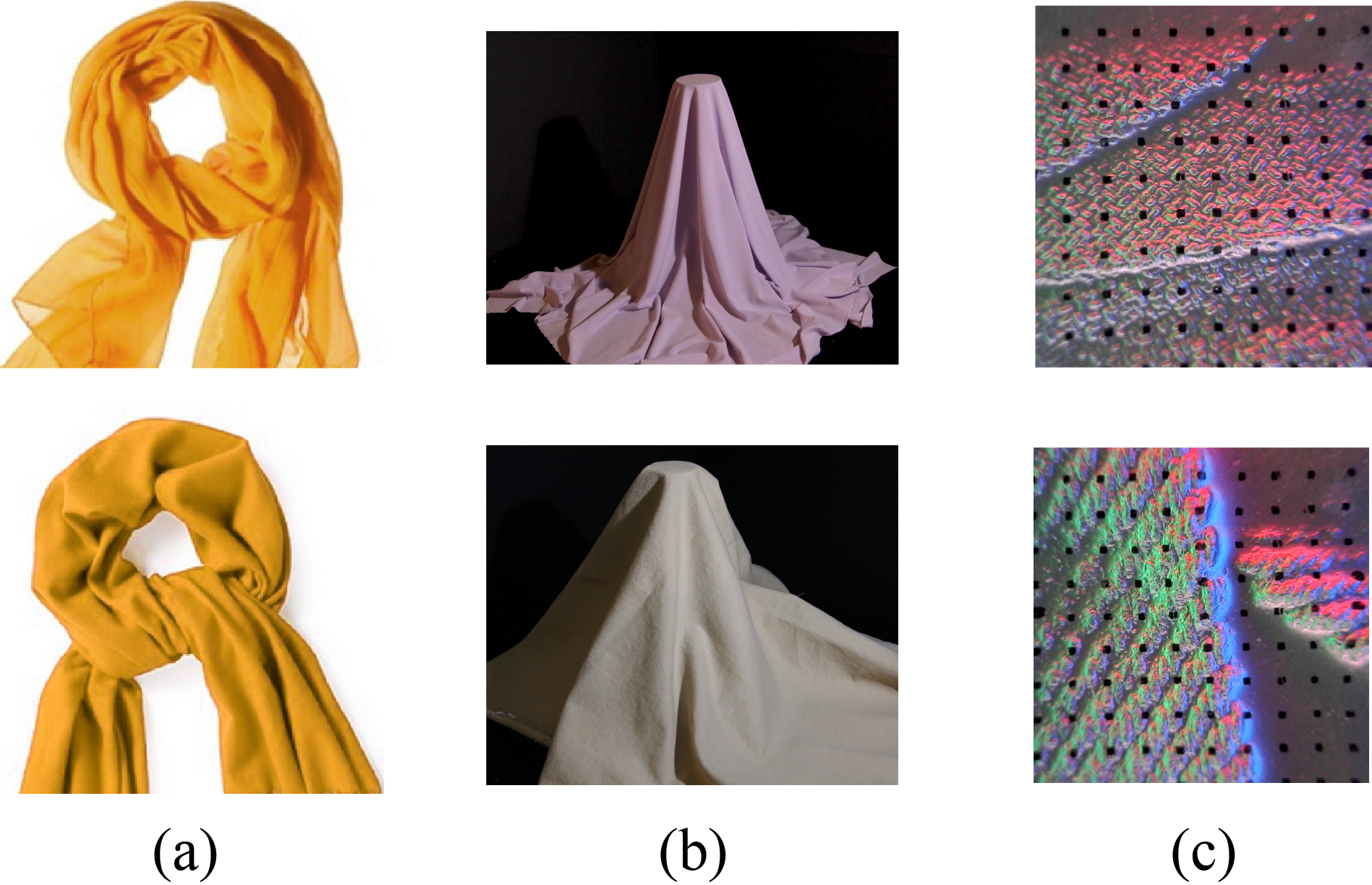}
	}
	\caption{Humans can infer fabric properties from draperies. 
		(a) A silk scarf and a wool scarf in similar configurations.
		(b) pictures of different fabrics draping from a cylinder post; (c) tactile images from GelSight when pressing on the fabrics in natural states. }
	\label{fig:Intro}
\end{figure}

Figure~\ref{fig:Intro}(a) shows a silk scarf, and a wool scarf in a similar configuration. The silk scarf is lighter, thinner, and more flexible, while the wool scarf is heavier, thicker, and stiffer. A human observer, viewing the images, can easily see the difference. In addition, one can imagine touching and grasping the two scarves; they will feel different when touched and will deform differently when grasped.

%

A piece of fabric has certain mechanical parameters. At the macro level, these can include, for example, the Kawabata values that describe bending, stretching, compression, roughness, and so on. 
The fabric’s mechanical parameters cause it to take on certain shapes in response to manipulation. Those shapes lead to the observed images. 
At the same time, the fabric’s mechanical parameters engender a certain range of tactile interactions. When touched, the silk scarf will feel smooth and easily deformed; the wool scarf will feel quite different. 

We can think of the fabric’s physical parameters as latent variables that influence the processes leading to both the visual and tactile signals. To humans, this quantified set of physical parameters better represents a fabric-- a human seldom infers to a fabric as a particular individual, but the parameter set, like ``the light, thin and flexible silk''. The same piece of fabric, may make different appearances or feelings to touch, but is still considered the same one, because they share the same set of parameters. On the other hand, some fabrics may be considered similar because their parameters have close values, while the other fabrics are distinct because their parameters are distant.

Those underlying parameters are never directly observed. Instead, they manifest themselves indirectly by yielding a certain range of sensory data.
The end result is that the way a fabric looks is related to the way it feels, even though the process of image formation is completely different from the process of tactile interaction. 

In this paper, we separately generate visual and tactile data for a given fabric, and try to learn the association through an auto-generated embedding vector, which should only related to the fabric's physical parameters. First, we drape the fabric over a cylindrical post to see the shapes that it forms. Each time we repeat the draping we get a somewhat different shape. 
Second, we touch the fabric with a high-resolution touch sensor GelSight, which captures fine details of the fabric surface. We press on a fold of the fabric so as to gather information about its thickness and stiffness as well as its texture.
For a piece of fabric, regardless of the sensory modality and the occasional appearance, the embedding vector is expected to be the same.

\begin{figure}
	\centering{
		\includegraphics[]{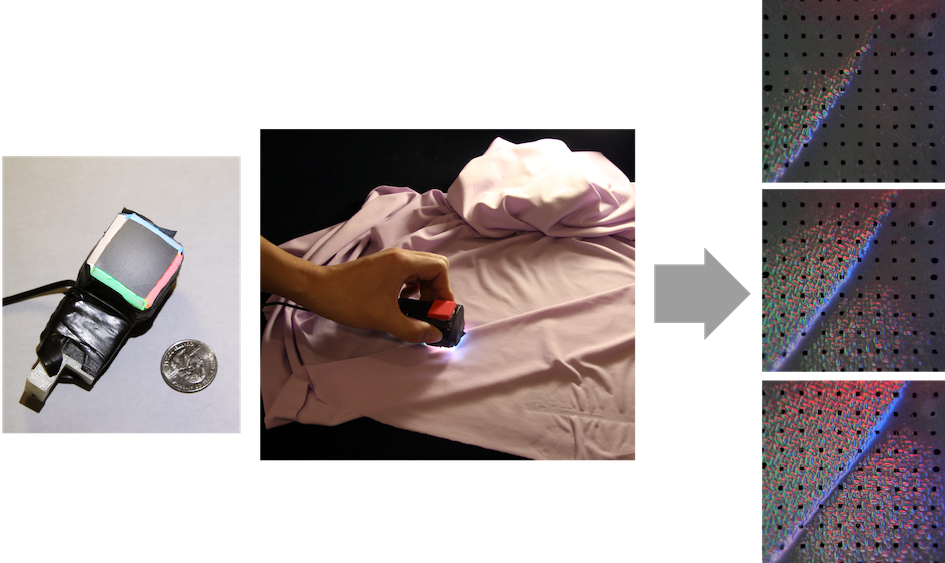}
	}
	\caption{Left: GelSight sensor. Middle: a human presses GelSight on the flat fold of a fabric, and gets a sequence of tactile images (shown in the right).}
	\label{fig:GelSight}
\end{figure}

\begin{figure}[t]
	\centering{
		\includegraphics[]{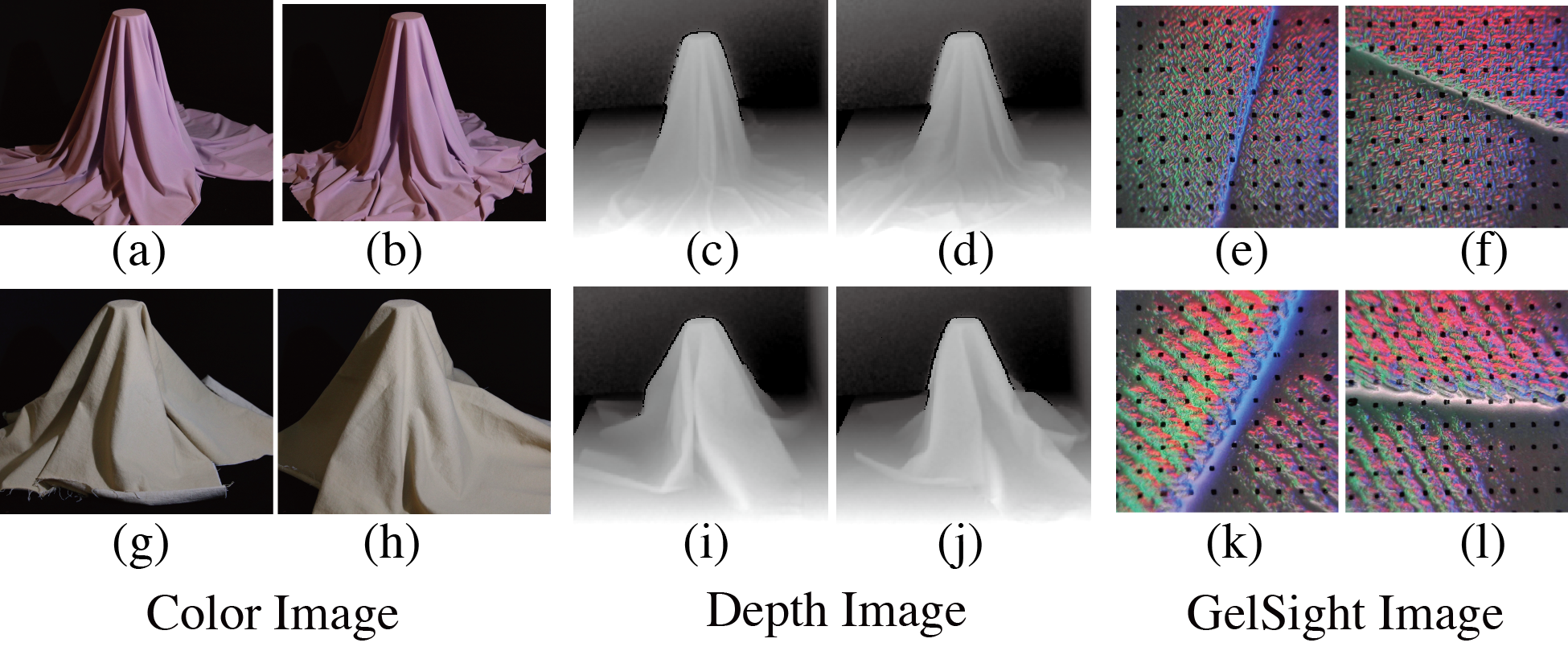}
	}
	\caption{Three modalities of the fabric data. For the visual information, the fabrics are draped from a cylinder in natural state; for the tactile information, a human holds the GelSight sensor and presses on a fold on the fabric. }
	\label{fig:DataExample}
\end{figure}

\begin{figure*}
	\centering{
		\includegraphics[scale=1.02]{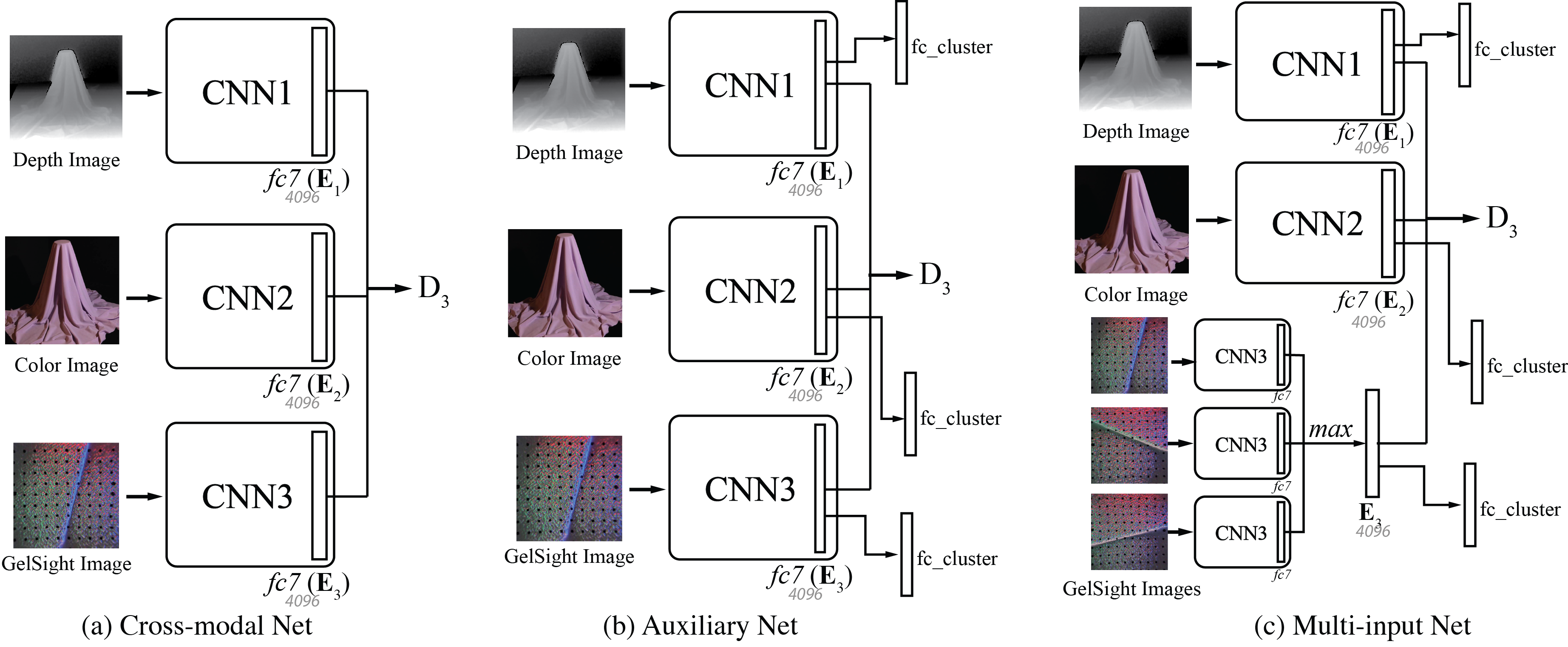}
	}
	\caption{
		The joint neural network architectures in this paper. Three modalities: depth image, color image, and touch image (GelSight image), are associated.
		(a) The Cross-modal Net: data from the three modalities goes through three independent CNNs (AlexNet~\cite{krizhevsky2012imagenet}) in a joint network, and be presented by an embedding $\mathbf{E}$, which is the $fc7$ layer of the network. We then compare the distance $D_3=\|\mathbf{E}_1-\mathbf{E}_2\| +\|\mathbf{E}_1-\mathbf{E}_3\| +\|\mathbf{E}_3-\mathbf{E}_2\|$ between the 3 embeddings.		
		For the same fabric, the three embeddings should be close and $D_3$ should be small.
		(b) The Auxiliary Network with the subtask of fabric classification. Clusters of the fabrics' are made according to human label. 
		(c) The Multi-input Network, that touch embedding is derived from 3 independent GelSight pressing images. }
	\label{fig:Network1}
\end{figure*}

Figure~\ref{fig:GelSight}(left) shows a GelSight sensor~\cite{GelSight2009,GelSight2011}. It employs a slab of clear elastomer covered with a reflective membrane. The device contains a small camera and lighting system, and uses optical methods to measure the deformation of the membrane as it is pressed against surfaces. For the present experiments, we press the sensor against a piece of fabric that is lying on a table, as shown in Figure~\ref{fig:GelSight}(mid). We intentionally introduce a fold in the fabric so that the GelSight images can show how the fabric bends when pressed. A sequence of output images are shown in Figure~\ref{fig:GelSight}(right). The colors correspond to the varying surface normal. The grid of dots, which is printed on the membrane, distorts in response to contact forces. 

To gather images of fabrics', we chose a standard arrangement that would reveal some mechanical properties. Figure~\ref{fig:DataExample}(a) shows an RGB image of a fabric draped over a cylinder. Figure~\ref{fig:DataExample}(b) shows the same fabric, draped a second time. The images are different, but both convey the mechanical properties of this fabric. Depth images, which isolate shape without the optical complications, are shown in Figure~\ref{fig:DataExample}(c) and (d). 
Figure~\ref{fig:DataExample}(e) and (f) show the output of a GelSight sensor when it is pressed (by hand) against a sample of the same fabric.  Each image is captured with the fabric, and its fold, in a different position and orientation.
Figure~\ref{fig:DataExample}(g) -(l) show the same images for a different sample of fabric. The second fabric is heavier and thicker than the first, and this leads to different appearances, in both the visual and the tactile domain.

We obtained a collection of 100 fabric samples for training, and 18 fabric samples for testing. For each fabric sample, we draped it over the cylindrical post for 10 times, producing 10 color and depth images. For each sample, we also generated 10 tactile images using the GelSight sensor. 

Our main task is this: given two input images -- either color, depth of tactile images, determine whether they were generated by the same fabric.
We design multiple neural network architectures for the cross-modality training and compare their performance.
We also compare the recognition performance on a single modality when joint trained on one or two modalities, and find that the extra information from another modality will boost the performance on single-modality match.


\section{Related work}

\textbf{Visual Perception on Fabrics}
Studies like ~\cite{Ted2001seeing,fleming2014visual} showed humans use visual cues to infer different material properties. 
Specifically, Xiao \etal.~\cite{xiao2016can} studied human perception on fabrics and the influencing factors by using tactile sensing as ground truth to measure visual material perception. They showed that humans made a high matching accuracy, while the color and 3D folds of the fabrics' are the most important to the human visual perception.

Researchers in computer vision and graphics have been trying to track or represent fabrics or clothes, but their visual representation is difficult to obtain compared with rigid objects', and the uncertainty and complexity of the shapes and motion make the fabrics or clothes more difficult to predict. 
To track the exact shape of the clothes, White \etal~\cite{CaptureFold2007} made dense patterns on fabrics or clothes, and used multiple cameras to track their motion thus to reconstruct the 3D shapes of the clothes. 
Han \etal~\cite{han2007Drapery} represented the cloth shape with a 2-layer model: one represents the general shape, the other one represents fold shapes, which are measured by shape-from-shading methods.
Some other researches tried to represent the fabrics by physical parameters, and estimated the parameters from the visual appearance. 
Baht \etal~\cite{CMU2003MotionSimulation} used a model made of physical properties, including density, bend stiffness, stretch stiffness, damping resistance and friction,  to describe and simulate clothes. They estimate the properties by comparing the real clothes' motion video with the simulated videos. 
Bouman \etal~\cite{KatieMotion} measured fabric properties (stiffness and density) directly from the video of fabric motion using hand-crafted features, when the fabrics were hanged and exposed to different winds.   

\textbf{Touch Perception with GelSight}
As the best high-resolution tactile sensor, GelSight is good at measuring the fine structure of object surfaces. The tactile signal from the sensor comes in the form of image, thus some typical computer vision algorithms could be applied on processing the signal. Li and Adelson~\cite{GelSightTexture} showed GelSight worked well on recognizing materials classes according to their texture. Moreover, GelSight also showed potentials to obtain physical properties of objects through physical interaction with the objects. Yuan~\etal~\cite{HardnessIROS,HardnessICRA} pressed GelSight on soft objects, and estimate the objects' hardness from the image sequences from GelSight. 

\textbf{Joint Neural Network}
Joint neural network is the network architecture that joins two or more separate networks for different inputs. 
Chopra \etal~\cite{chopra2005learning} first proposed a Siamese Neural Network (SNN), that learned low dimensional embedding vectors from a single-modal input. The SNN has two identical neural networks with shared weights, and outputs the distance of embedding vectors from the two inputs. In the training, the network uses energy-based contrastive loss~\cite{hadsell2006dimensionality} to minimize the distance of the embeddings from similar input pairs while making the distance of dissimilar input pairs' embeddings larger than margin. SNN has been applied in face verification~\cite{chopra2005learning} and sentence embedding~\cite{mueller2016siamese}.


In recent years, people have been using joint neural network for cross-modality learning -- mostly 2 modalities.
A traditional method is to extract features from one modality and project the other modality to this feature space.
Frome \etal~\cite{frome2013devise} proposed hinge rank loss to transform visual data to text. Li \etal~\cite{li2015joint} learned the joint embedding by associating generated images to the trained embeddings from shape images of objects.
Owens \etal~\cite{AndrewSound} combined CNN and LSTM to predict objects’ hitting sound from videos. They extracted the sound features first, and then regress the features from images by neural networks. Their other work~\cite{owens2016ambient} presented a CNN that learn visual representation self-supervised by features extracted from ambient sound. 

\begin{figure*}[t]
	\centering{
		\includegraphics[]{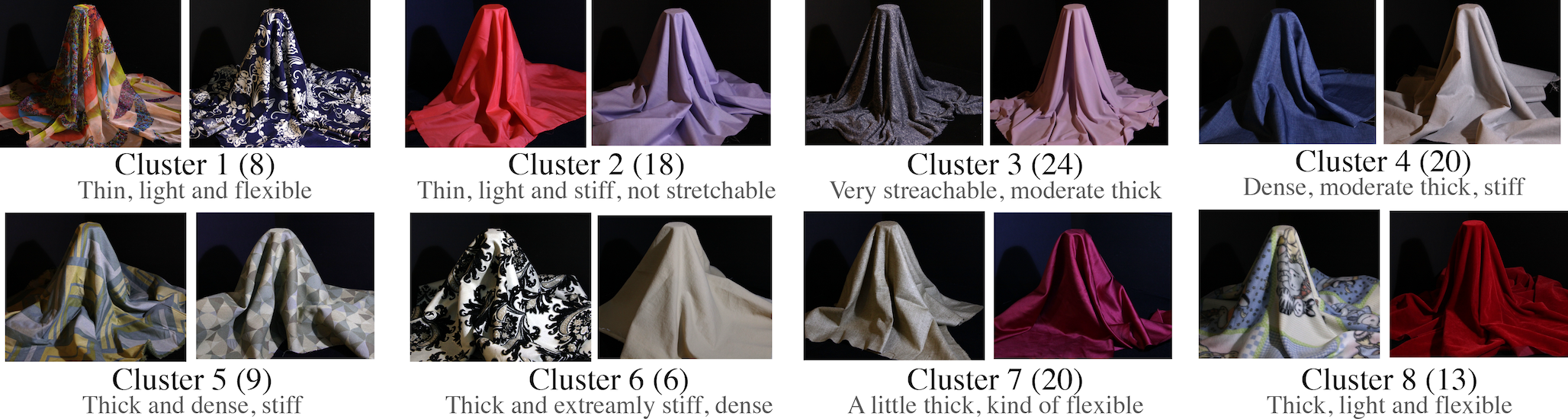}
	}
	\caption{Clustering of the fabrics based on human label. Numbers in the bracket denote the fabric number in the cluster. }
	\label{fig:Cluster}
\end{figure*}

Some other recent works have been trying to project inputs from different modalities into a shared embedding space.
Otani \etal~\cite{otani2016learning} proposed a network that learns cross-modal representations of videos and sentences simultaneously. Besides these bi-modal networks, Aytar \etal~\cite{aytar2016cross} proposed multi-modal neural networks that learn cross-modal representation of more than two modalities related to scenes. Projecting more modality to a shared space makes the learning process more difficult, but could bring more information. 

\section{Dataset}

We collect a dataset for fabric perception that consist of visual images (color and depth), GelSight videos, and human labelling of the properties. The dataset contains 118 fabrics, including the apparel fabrics like broadcloth, polyester, knit, satin; bedding fabrics like terry, fleece; and functional fabrics like burlap, curtain cloth, oilcloth (examples shown in Figure~\ref{fig:DataExample}). About 60\% of the fabrics are of single but different colors, others have random color patterns. Each piece of the fabrics is of the approximate size 1m$\times$1m. Some of the fabrics are kindly provided by researchers working on \cite{xiao2016can} and \cite{KatieMotion}. 
The dataset is available at \url{http://people.csail.mit.edu/yuan_wz/fabricdata/GelFabric.tar.gz} .

\textbf{Visual data} 
We drape the fabrics from a cylindrical post(30.7cm height, 6.3cm diameter) in natural states, and take both the color images and depth images of them. 
The color images are taken by a Canon T2i SLR camera, and depth images are taken by a Kinect One. For each fabric we take pictures of 10 different drapes.

\textbf{Tactile data} 
We press the tactile sensor, GelSight, on the fabrics when they lay on a hard flat surface, thus obtaining a sequence of GelSight tactile images for the press process. 
The sensor we used is the fingertip GelSight device~\cite{GelSightUSB}. The sensor has a slightly domed surface and a view range of 18.4mm$\times$13.8mm. We select the image resolution of $960\times 720$. There are black markers painted on the sensor surface to track the contact force~\cite{GelSightShear}.

We collected two forms of tactile data: one is the ``flat data'', when the GelSight is pressed on the single-layer of the flat fabrics; the other one is ``fold data'', when the GelSight is pressed on the fold of the fabrics, as shown in Figure~\ref{fig:GelSight}. 
For each fabrics, we collect 10 pressing samples of the flat data and 15 samples of the fold data.

\textbf{Attribute label}
We label each fabrics with the estimation of the physical parameters that we believe are the most important determine the fabric draping and contact process: \textit{thickness, stiffness, stretchiness} and \textit{density}.
 The thickness and density are measured by a ruler and a scale; stretchiness is roughly estimated into the level of ``non-stretchable'', ``stretchable'', and ``extremely stretchable''; the stiffness is estimated by humans: we ask 5 human subjects to score the fabric stiffness in the range of 0 to 5 (with the permission of excess for extra stiffness), and take the mean value. 
Note that the label does not necessarily cover all the true properties that influence the drape, and the values contains human bias, but they can provide a convenient and reasonable reference. 

In this work, we cluster the fabrics into 8 clusters by using k-means on the fabrics' physical parameters, as shown in Figure~\ref{fig:Cluster}. 
To humans, fabrics in the same cluster will have relatively similar properties. We describe the human intuitive description of each cluster in Figure~\ref{fig:Cluster}. 

\section{Associating Vision and Touch}

We build joint neural network models to associate visual and tactile information of the fabrics. The input data is of three different modalities: the depth image, the color images, and the tactile images from GelSight.
The input data from each modality goes through an independent CNN to form an embedding vector $\mathbf{E}$, as a low-dimension representation of the fabrics. We use the sum of Euclidean distance $D=\|\mathbf{E}_1-\mathbf{E}_2\|$ to measure the differences between two $\mathbf{E}$s, regardless of the input's modality. 
Ideally all the input data on the same fabric will make the same $\mathbf{E}$ through the networks, while two fabrics, when they are similar, will have a small distance $D$ between the embedding vectors $\mathbf{E}$, and two very different fabrics will have large $D$.
We trained a joint CNN of the three modalities, and compared the performance of different architectures.
Figure~\ref{fig:Network1} shows the neural networks in this paper.

\begin{figure*}
	\centering{
		\includegraphics[]{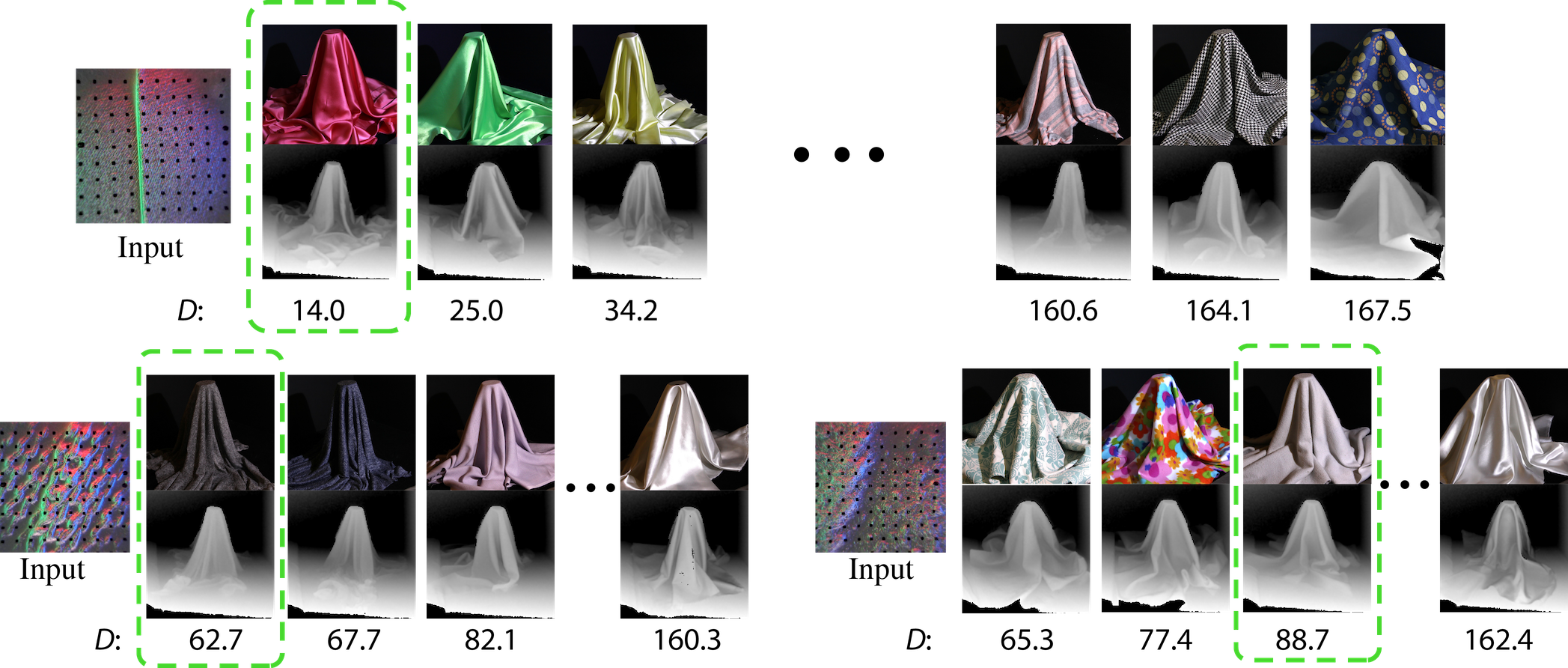}
	}
	\caption{Examples of picking the corresponding depth image to the GelSight input,  according to the distance $D$ between their embeddings. Trained on the Auxiliary Net. The green frames mark the ground truth.
		The first row shows an example of in training: a flexible and thin satin, where the 3 closest matches are all satins, and the furthest 3 examples are all different fabrics. 
		The second row shows two test examples, with the 3 closest matches and a furthest match in the random 10-image set. The right example shows a soft, light and flexible fleece, but the network confused the sample with two other thick and soft blankets. }
	\label{fig:PickExample}
\end{figure*}

\subsection{Neural Network Architectures}

\textbf{Cross-modal Net}
The basic network to join the three modalities is shown in Figure~\ref{fig:Network1}(a). In this network, the architecture images, color images and GelSight images go through three separate CNNs in a joint network. The CNN we used in this work is the AlexNet~\cite{krizhevsky2012imagenet}, which is pretrained on ImageNet, and we take the $fc7$ in the network as the embedding vector $\mathbf{E}$ to represent a fabric.

We use contrastive loss\cite{chopra2005learning} as objective function. For a input group of depth image $X_1$, color image $X_2$ and GelSight image $X_3$, the embedding vectors coming from the three neural network $G_{W1}$, $G_{W2}$ and $G_{W3}$ can be denoted as $\mathbf{E}_1 = G_{W1}(X_1)$, $\mathbf{E}_2 = G_{W2}(X_2)$ and $\mathbf{E}_3 = G_{W3}(X_3)$. For each input group, we measure the overall distance between the embedding vectors, denoted as $D_3$:
\begin{equation}
D_3=\|\mathbf{E}_1-\mathbf{E}_2\| +\|\mathbf{E}_2-\mathbf{E}_3\| +\|\mathbf{E}_3-\mathbf{E}_1\|
\end{equation}
We make $Y = 0$ if $X_1$, $X_2$ and $X_3$ are sourced from the same fabric, and $Y = 1$ if they are from different fabrics. The network loss is 
\begin{equation}
\begin{split}
L(W1, W2, Y, X_1, X_2) = & \frac{1}{2}(1-Y) \times D_3^2 + \\ 
& \frac{1}{2}Y\times {\max(0,m-D_3)}^2
\end{split}	
\label{eq:Loss}
\end{equation}
where $m>0$ is a margin (we used $m=2$ in our experiments). Dissimilar pairs contribute to the loss function only if $D_3$ is smaller than the margin radius $m$. The existence of dissimilar pairs are meaningful to prevent the $D_3$ and the loss $L$ being zero by setting $G_{W}$s to a constant.

\textbf{Auxiliary Net}
In auxiliary net, we keep the architecture of the basic cross-modal net, but simultaneously use the embedding vector $\mathbf{E}$ to train a classification task of the fabrics cluster, as shown in Figure~\ref{fig:Network1}(b). The purpose is to make similar fabrics have close embedding vectors by adding supervision. The three cross-entropy losses of cluster classification are combined with the contrastive loss(\ref{eq:Loss}) in addition for a total loss.
The cluster of the fabrics is made based on human label, as shown in Figure~\ref{fig:Cluster}. 

\textbf{Multi-input Net}
Based on the auxiliary network, we use 3 different GelSight images different presses as tactile inputs, thus making the Multi-input Network. The 3 GelSight images go through the same network $G_{W3}$ respectively, making 3 $fc7$ vectors, and we make the final embedding $\mathbf{E}$ of the inputs as element-wise maximum of them. 
The network is shown in Figure~\ref{fig:Network1}(c). 
The motivation for this design is that, humans are likely to touch an object for multiple times before obtaining a confident perception of it, and similarly, we design the multi-input architecture to exploit more information from the multiple presses. 

\subsection{Training and Test}

Our approach is implemented in Keras\cite{chollet2015keras} with TensorFlow\cite{tensorflow2015-whitepaper} backend. We use the Adam \cite{kingma2014adam} optimizer and fix learning rate as 0.001 throughout the experiment. Parameters of AlexNet before $fc7$ will be fixed during training. We train the network for 25,000 iterations with $batch\_size=128$. 

In the test, we used the trained CNNs $G_{W1}$, $G_{W2}$ and $G_{W3}$. Each input image, either a depth image, color image or GelSight image, goes through the corresponding network to produce an embedding $\mathbf{E}$, as a representation of the fabric. For different inputs, either from the same or different modalities, we calculate the $\mathbf{E}$s from the input, and compare the distance $D$ between the two $\mathbf{E}$ to decide the likeliness that the two inputs are from the same fabric.

\section{Experiments}

We divide the 118 fabrics in the dataset as a training set (100 fabrics) and test set (18 fabrics). The 18 test fabrics are selected evenly from the 8 clusters in Figure~\ref{fig:Cluster}. 

\subsection{Infer Touch from Vision}
\label{chapt:exp1}

The first experiment is picking the depth or color images that best match the GelSight input. In other words, we give the network a touch image, and some possible visual appearances of the fabrics, then we ask the network to choose the most probable image of the touched fabric. The match is according to the $D$ between the $\mathbf{E}$s from the given GelSight image and the candidate images. 
In the experiment, the candidate depth or color images are 10 images from 9 random selected fabrics and the ground-truth fabric from the test set. 
The selecting procedure is shown in Figure~\ref{fig:PickExample}.
We evaluate the model performance by comparing the top 1 precision and top 3 precision: the probability of the correct answer ranks the first in all the 10 candidates, or ranks in the top 3.
For each network, we test each 15 different GelSight input images on each fabric for 10 times, and calculate the average precisions.

\begin{figure}
	\centering{
		\includegraphics[width=\linewidth]{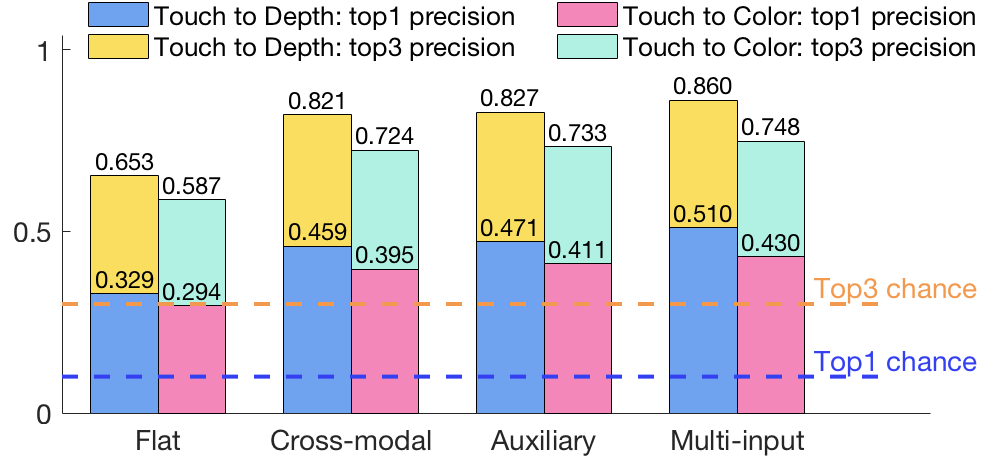}
	}
	\caption{Test result: the top 1 and top 3 precision on matching the depth or color image candidates to a given GelSight input. We compared 4 models: the Cross-modal net with input GelSight images of pressing on flat fabrics (denoted as `Flat'), the Cross-modal net, the auxiliary net and the Multi-input Net. (the last three results groups are based on GelSight images when pressing on folded fabrics) }
	\label{fig:PickRes}
\end{figure}

We test the performance of 4 networks: 1. the cross-modal network (Figure~\ref{fig:Network1}(a)), when the GelSight input is the pressing image on flat fabrics without folds;
2. the cross-modal network, when the GelSight input is one pressing image on the folded fabrics;
3. the auxiliary network (Figure~\ref{fig:Network1}(b)) that compares depth images and GelSight on single folds, but with the auxiliary task of cluster the embeddings; 
4. the auxiliary network that takes 3 GelSight images as the input (Figure~\ref{fig:Network1}(c)).
The results of the top 1 precision and top 3 precision on the test set is shown in Figure~\ref{fig:PickRes}. 
We also test the precisions of matching other modalities, and the results are shown in Table~\ref{Tb:OtherMatch}.
In comparison, the precisions on matching the data from a single modality is much higher, as shown in Tabel~\ref{Tb:OneModalMatch}. 

From the results, we can see that all the networks can predict the matching images better than average chance. As for the architectures, the auxiliary net with 3-frame input performs the best, the auxiliary net with 1-frame input places the second, and the basic model with the plain GelSight press comes the last. 
The match between touch images and depth images is better than the match with color images.

\begin{table}
	\begin{center}
		\begin{tabular}{p{1.6cm}|c|c|c|c}
			\hline
			Model & Flat & Cross-mdl & Auxiliary & Multi-in\\
			\hline\hline
			Depth2Gel & 0.3063 & 0.4292 & 0.4318 & 0.4576\\
			Color2Gel & 0.2681  & 0.3742 & 0.4022 & 0.4124\\
			Depth2Color & 0.4133 & 0.4329 & 0.4141 & 0.4417\\
			Color2Depth & 0.4050 & 0.4240 & 0.4070 & 0.4306\\
			\hline
		\end{tabular}
	\end{center}
	\caption{Result on the test set: the average top 1 precision on test set for the ``pick 1 from 10'' experiment of matching other modalities.}
	\label{Tb:OtherMatch}
\end{table}

\begin{table}
	\begin{center}
		\begin{tabular}{p{1.6cm}|c|c|c|c}
			\hline
			Model & Flat & Cross-mdl & Auxiliary & Multi-in\\
			\hline\hline
			Dep2Dep  & 0.6030 & 0.6265 & 0.6224 & 0.6459\\
			Color2Color & 0.7941 & 0.7831 & 0.7968 & 0.8247\\
			Gel2Gel & 0.8025 & 0.7672 & 0.8090 & 0.9351\\
			\hline
		\end{tabular}
	\end{center}
	\caption{Result on the test set: the average top 1 precision on test set for the ``pick 1 from 10'' experiment of matching a single modality.}
	\label{Tb:OneModalMatch}
\end{table}

\begin{figure*}[ht]
	\centering{
		\includegraphics[]{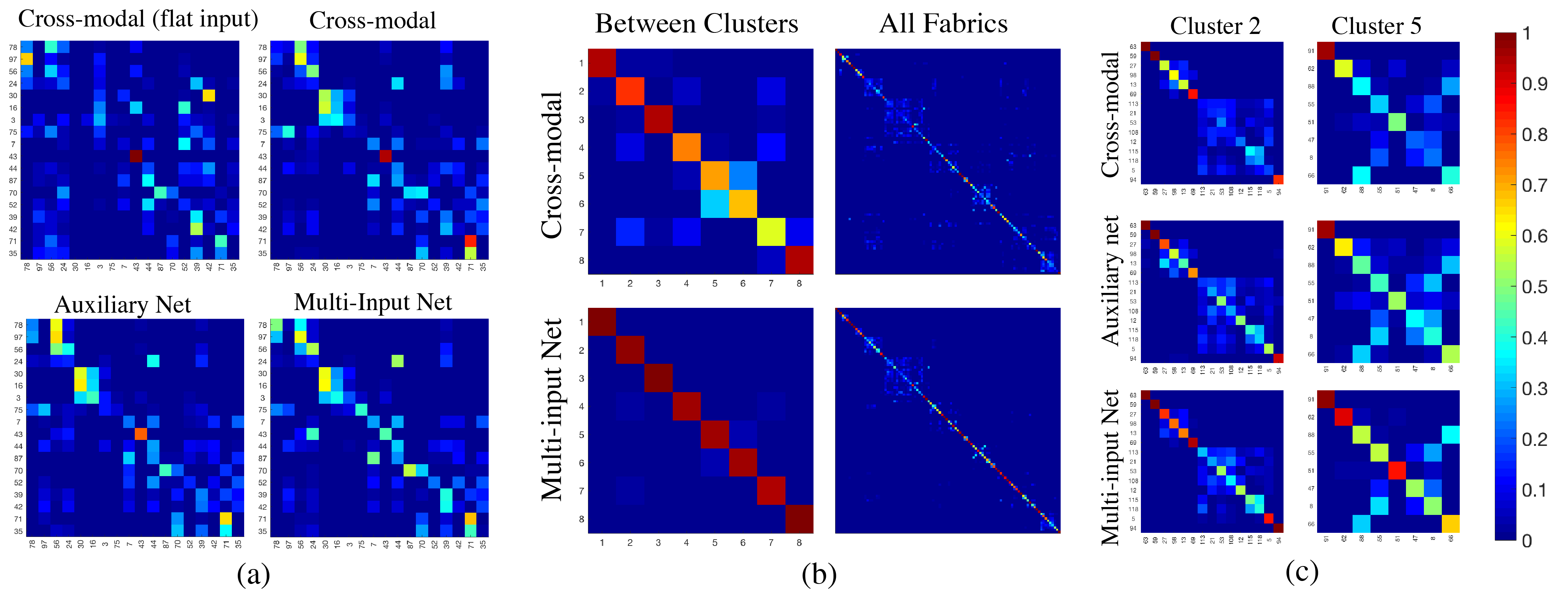}		
	}
	\caption{Confusion matrices between fabrics on ``picking the possible depth image to a given GelSight input''. The fabrics are ranked according to human subjects, so that similar fabrics are placed close.
		(a) Confusion matrices of the test results for different networks.
		(b) Confusion matrices on training set for the Cross-modal net and Mutli-input Net, either between clusters, or on the individual fabrics.
		(c) Confusion matrices on fabrics in the training set within Cluster 2 and Cluster 5.}
	\label{fig:ConfMatRes}
\end{figure*}

The positive results in the matching experiments show that the neural networks are able to automatically extract the features related to fabric intrinsic properties from either visual or touch information. The properties from the three modalities are correlated, so that the networks can match one modal input with the other by comparing the embedding vectors. 
But in the given dataset with the limited size, the neural networks extract the physical properties better from the depth images than from the color images, because the former has less information and the fabric shape is more directly related to the physical properties.
The results also show that, the additional information helps the network to better recognize the materials: the comparison between model 1 and 2 shows that the folds on the fabric reveals more properties; comparison between model 2 and 3 shows on this small dataset, the human label help to improve the network performance; the comparison between model 3 and 4 shows that providing more touch information, the network will extract the relevant information better, and makes the matching more robust.

\subsection{Representing Fabrics by Embeddings}
\label{chpt:Exp2}

For each input image, we represent it with an embedding vector $\mathbf{E}$ through the trained neural network. The distance $D$ between two  $\mathbf{E}$s is expected to measure the likeliness that they are sourced from the same fabric, or two similar fabrics. In this experiment, we aim at seeing how the $\mathbf{E}$ represents the fabrics; in other words, whether $\mathbf{E}$s from the same or similar fabrics are closer than those of distinct fabrics.

In this section, we continue with the experiment of ``picking the possible depth image given a GelSight image'' as an example. To denote the possibility that the two $\mathbf{E}$s are sourced from the same fabric, we build a function $P$:
\begin{equation}
\mathrm{P}(\mathbf{E}_1, \mathbf{E}_2) \propto \exp \left( -c\times D(\mathbf{E}_1, \mathbf{E}_2)^2 \right)
\end{equation}
Where $c$ is a positive coefficient (we set it as $8.5\times 10^{-2}$). For a given input with embedding $\mathbf{E}_{tar}$, and a set of candidates with embeddings $\left\{\mathbf{E}_i \right\}$, we normalized $\mathrm{P}$ so that 
\begin{equation}
\displaystyle\sum_{i}\mathrm{P}(\mathbf{E}_{tar}, \mathbf{E}_i)=1
\end{equation}
Here we make $\left\{\mathbf{E}_i \right\}$ from all the depth images in the candidate fabric set. For each test fabric, we calculate $P$ over all the available GelSight input image and take their average, so that we got a possibility of ``mismatching the touch data from the current fabrics to the other fabrics''.
We draw confusion matrices of the mean $\mathrm{P}$ between the fabrics in Figure~\ref{fig:ConfMatRes}. In the figure, we re-order the fabrics numbers to put the fabrics adjacent when human subjects consider them similar, so that the bright spots near the diagonal line means the neural network gets confused with the fabrics that are likely to confuse human too.

Figure~\ref{fig:ConfMatRes}(a) shows the confusion matrix on the test dataset, and it indicates that most of the possible confusion occurs between the similar fabrics. 
For instance, in the examples shown in Figure~\ref{fig:PickExample}, The first test case, the network picked the correct candidate, but the $D$ is close to the second candidate, because the two fabrics are similar knits; in the second text case, the network predicted wrong, because the input pattern is like a ``thick, soft and fuzzy material'', while the best matched and second match have similar properties. 
In general, the Multi-input Net performs the best on the confusion distribution, while the Cross-modal Net with only plain input performs the worst.

Figure~\ref{fig:ConfMatRes}(b) shows the probability in matching the GelSight data and depth image in the training set (100 fabrics). Here we compared the matching probability of all the independent fabrics, and also between different clusters. 
The figures indicate that for both networks, they well distinguish the fabrics in different clusters. Even the Cross-modal net does well, while it does not know the cluster in the training. But with in the clusters, the network can be confused between fabrics. 
Figure~\ref{fig:ConfMatRes}(c) shows the confusion matrices of fabrics within Cluster 2 and 5. 
Cluster 2 denotes fabrics that are ``thin, light and stiff'', and contains many broadcloths. They appear very similar to human; similarly, the Cross-modal Net and Auxiliary Net make their embedding vectors close, and display a blurred area in the bottom left part in the matrices.
But for the Multi-input Net, as there are more input information, the network is able to represent the more subtle differences between the fabrics, so that the confusion matrix concentrated. 
Cluster 5 contains fabrics that are thick and stiff. Similarly, the Multi-input Net reduced the confusion between different fabrics the best (although not totally), and the embedding vectors would better represent the fabrics.  

The results in this section prove that all those factors will improve the network's ability of representing the fabrics: touching the folds instead of the plain fabric; multiple presses that contains less biased information. The clustering information made according to human label also help the network to narrow down the fabric range to represent the properties.

\subsection{Data Augmentation}

We augment the dataset on the color images by changing the hue and exposure of the images during the training: we performs Gamma Correction (range 0.5-2.0) to each image, and change the order of the RGB channels. The matching tests with the color images involved make a better result, as shown in Table~\ref{Tb:ColorAug}. But the results of other matching tests between GelSight images and depth images do not change. 
We tried other data augmentation on the GelSight images and the depth images, including adding noise to the input,  and crop the images randomly, but the results make little difference.

\subsection{Touch Helps Vision}

We find that the joint learning of the multi-modality boosts the performance on one modality. Taking the vision as an example, we work on the task of ``picking a depth image of draped fabrics that best matches a given depth image''. 
We compare the performance of 2 network architectures: a Siamese Neural Network (SNN)~\cite{chopra2005learning} trained on only depth images, and a Cross-modal Net similar to Figure~\ref{fig:Network1}(a), but only takes in depth images and GelSight images. 
The two architectures are the same other than they take in different modalities as branches.  

\begin{table}
	\begin{center}
		\begin{tabular}{p{1.6cm}|p{1.47cm}|p{1.47cm}|p{1.18cm}|p{1.45cm}}
			\hline
			Model & Cross-mdl & Cross-mdl (with aug) & Multi-in & Multi-in (with aug)\\
			\hline\hline
			Gel2Color & 0.3954 & 0.4359 & 0.4303 & 0.4937\\
			Color2Gel & 0.3742 & 0.4088 & 0.4124 & 0.4264\\
			Depth2Color & 0.4329 & 0.4674 & 0.4417 & 0.4924\\
			Color2Depth & 0.4240 & 0.4607 & 0.4306 & 0.4624\\
			\hline
		\end{tabular}
	\end{center}
	\caption{Comparison of the top 1 precision before and after data augmentation on the color images.}
	\label{Tb:ColorAug}
\end{table}

\begin{table}
	\begin{center}
		\begin{tabular}{c|c|c|c|c}
			\hline
			 & \multicolumn{2}{ |c| }{Seen Fabrics}  &\multicolumn{2}{ |c }{Novel Fabrics} \\
			\cline{2-5}
			Model & Top1 & Top3 & Top1 & Top3\\
			\hline\hline
			SNN (only depth) & 0.482 & 0.660 & 0.554 & 0.729\\
			Cross-mdl (depth\&Gel)& 0.608 & 0.786 & 0.606 & 0.786\\
			\hline
		\end{tabular}
	\end{center}
	\caption{	Test results on the depth-to-depth match on two networks: a Siamese Neural Network (SNN)~\cite{chopra2005learning} trained only on depth images, and a Cross-modal Net trained on depth and GelSight images.}
	\label{Tb:VisionMat}
\end{table}

In this experiment, we select 80\% of the data on the 100 training fabrics as the training set, and the rest 20\% data, as well as the data from 18 test fabrics as the test set. The test results are shown in Table~\ref{Tb:VisionMat}.
As shown in the results, on this size-limited dataset, the joint model on both touch and depth images have much better performance than single-modal SNN model.
We assume this means the extra information from one modality will help the training in the other modality to reduce overfit and find a better local minimum.

%

\section{Conclusion}

In this work, we use deep learning to associate visual and tactile information on recognizing fabrics. Three modalities are used: the depth image, the color image, and the GelSight image. 
The recognition is more about estimating the physical parameters of the materials, rather than a discrete label, and the parameters are represented by an auto-trained embeddings. 
The distance between two embedding vectors shows how likely the data source are the same fabric, or how similar the fabrics are.
We compare the performance of different neural network architectures which exploit different amounts of input information, and the results show that, the folds on the fabrics during touching, the presumed fabric cluster based on human labels, and the multiple touch input, will help the network to learn better embedding vectors. 
The comparison of networks trained on single modality and two modalities also shows that, the joint training with both visual and tactile information will greatly improve the performance on visual information matching.

\begin{flushleft}
\textbf{Acknowledgment} 
 \end{flushleft}
The authors thank Andrew Owens, Bei Xiao, Katie Bouman and Tianfan Xue for helpful discussion and suggestions. The work is supported by Toyota, Shell, NTT, and NSF.\\

{\small
\bibliographystyle{ieee}
\bibliography{egbib}
}

\end{document}